\documentclass{article}

\usepackage{graphicx}
\usepackage{subfig}

\usepackage[preprint]{neurips_2021}

\usepackage[utf8]{inputenc} 
\usepackage[T1]{fontenc}    
\usepackage{hyperref}       
\usepackage{url}            
\usepackage{booktabs}       
\usepackage{amsfonts}       
\usepackage{nicefrac}       
\usepackage{microtype}      
\usepackage{xcolor}         

\title{Learning Stable Representations with Full Encoder
}

\author{%
  Zhouzheng Li\\
  Department of Mechanical Engineering\\
  Beijing University of Chemical Technology\\
  \texttt{Remdoeno@foxmail.com} \\
   \And
   Kun Feng \\
   Department of Mechanical Engineering \\
   Beijing University of Chemical Technology \\
   \texttt{kunfengphd@163.com} \\
}

\begin{document}

\maketitle

\begin{abstract}
While the $\beta$-VAE family is aiming to find disentangled representations and acquire human-interpretable generative factors – like what an ICA (from the linear domain) does, we propose Full Encoder – a novel unified autoencoder framework as a correspondence to PCA in the non-linear domain. The idea is to train an autoencoder with one latent variable first, then involve more latent variables progressively to refine the reconstruction results. The Full Encoder is also a latent variable predictive model that the latent variables acquired are stable and robust, as they always learn the same representation regardless of the network initial states. Full Encoder can be used to determine the degrees of freedom in a simple non-linear system and can be useful for data compression or anomaly detection. Full Encoder can also be combined with the $\beta$-VAE framework to sort out the importance of the generative factors, providing more insights for non-linear system analysis. These qualities will make FE useful for analyzing real-life industrial non-linear systems. To validate, we created a toy dataset with a custom-made non-linear system to test it and compare its properties to those of VAE and $\beta$-VAE’s.
\end{abstract}

\section{Introduction}
\label{Intro}
The Autoencoders (AE) have been used to automatically extract features from data without supervision for many years. Since then, a lot of work has been done to enhance this elegant structure. One particular enhancement direction focuses on extracting latent variables with properties that’s useful for non-linear system analysis: Contractive autoencoder [1] by Salah Rifai, et al, (2011) encourages the learned representation to stay in a contractive space for better robustness; Variational autoencoder [2] by Diederik P Kingma and Max Welling (2014) rooted in the methods of Variational Bayesian and graphical model, mapping the input into a distribution instead of individual variables; Diederik P. Kingma, et al, in the same year also introduced conditional VAE [3] to learn with labeled data and obtain meaningful latent variables with semi-supervised learning; then $\beta$-VAE [4, 5] is proposed by Irina Higgins, et al, (2017), trying to learn disentangled representations by strengthening the punishment of KL term with a hyperparameter beta and narrowing the information bottleneck, $\beta$-TC VAE [6] by TianQi Chen, et al, (2018) further refined their work until Babak Esmaeili, et al, (2019) unified all methods that modifies objective function with HFVAE [7].

From the GAN [8] family there are also methods that try to do something similar, a few milestones are conditional GAN (2014) [9] that involves label to input data, BiGAN (2017) [10] with a bidirectional structure to project data back to latent space, InfoGAN (2017) [11] that learns disentangled representations with information-theoretic extended GAN, and InfoGAN-CR (2020) [12] that references techniques from the $\beta$-VAE branch.

While the previous works mainly focus on disentanglement or interpretability of the representations: like what Independent Component Analysis~(ICA) can achieve in the linear domain, the field still lacks a correspondence for Principal Component Analysis~(PCA) in the non-linear domain. In this paper, Full Encoder (FE) is proposed as a combination of a few of the older AE models, along with a new Progressive Patching Decoder (PPD) structure to learn representations with another unique property: stability -- like in PCA where the principal components always stays the same. The representations learnt by FE are ranked by their importance for reconstruction, and for the same dataset FE always provides the same result, achieving stability. The FE framework can reflect the decreasing reconstruction error with more degree of freedom provided by the latent variables. Good robustness is achieved by introducing denoising structure [13] to FE using a dropout layer in the beginning, this allows the FE to perform a self-supervised learning task [14] that masks some observable variables from the system and let the network to learn a more general pattern to guess what the masked variable should be, therefore understanding the non-linear system better. The highly flexible structure also enables FE to work both on supervised or unsupervised manner, making this framework useful while dealing with real-life industrial scenarios.

A series experiments is conducted varing the amount of latent variables in FE to see its properities with a toy dataset sampled from a costume-made non-linear system. The results of FE is also compared with VAE and $\beta$-VAE model. We also conducted 3 supplementary expirments including one with digital number MNIST [15] to further explore FE's potentials. The results are presented in the Experiment Section and Appendix B from the supplemental material.

\section{Full Encoder}

\subsection{Overview}

Full Encoder is made of 2 parts – like a classical autoencoder – an encoder, and a decoder. The encoder of FE can be further divided into 2 neural nets, one for supervised label $y$ regression or categorization named \textbf{\emph{Encoder0}}, the other \textbf{\emph{Encoder}} for generating unsupervised latent variables $z$. The decoder of FE is a multi-level decoder with the name “Progressive Patching Decoder (PPD)”, it takes in not all the latent variable $z$’s but one at a time to progressively patch the "median code" $m_{i}$ with "process variable" $p_{i}$ generated from $z$ with neural nets named $NN_{i}$, then followed by a normal \textbf{\emph{Decoder}}to reconstruct original data $x$. The structure of Full Encoder is illustrated in Fig.~1.

\begin{figure}
  \centering
  \includegraphics[width=1.0\textwidth]{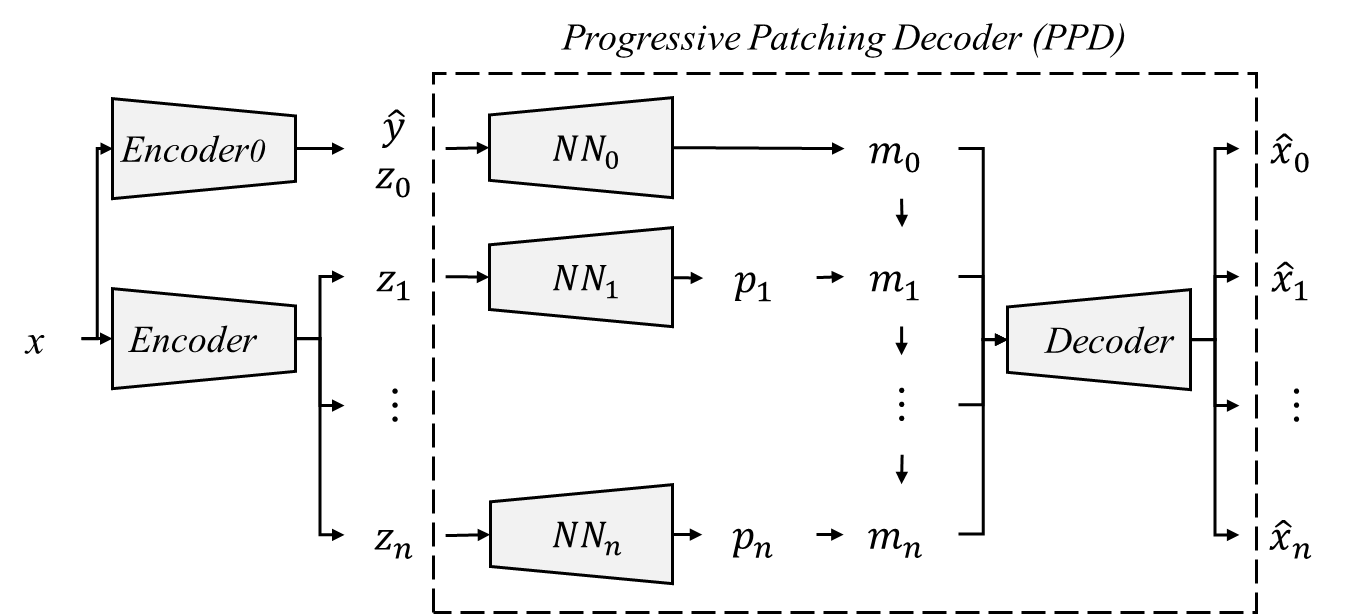}
  \caption{The complete structure of Full Encoder.}
\end{figure}

The main objectives for Full Encoder are: minimizing the mean square error (MSE) between input data vector $x$ and reconstructed data vector $\hat{x_{i}}$, minimizing the label loss if any label is provided, and minimizing the negative KL-divergence between prior and posterior distribution from VAE. What’s special about FE is that instead of having one reconstructed result, we have $n$ of them. Each of the reconstructed result $\hat{x_{i}}$ uses the first $i$ latent variables, from $z_{0}$ to $z_{i}$, as if we are training $n$ individual autoencoders with $1\sim n$ latent variables simultaneously with shared weights. This mechanism is how the regularization is done to FE to acquire latent variables that always learn the same pattern and achieve reconstruction results nearing a complete autoencoder. With the proposed PPD, we are able to merge the $n$ autoencoders together, and to fuse information provided by each latent variable with a unified shape.

\subsection{Encoder Network}

The encoder network does feature extraction for the Full Encoder. With labels provided, the structure is just like a mutation of CVAE [4]. While using without supervision, the structure is a standard VAE. The biggest difference between the FE encoder network and a standard CVAE or VAE encoder is its objective. Though no new terms have been added, modifications have been made to the loss function in collocation to the Progressive Patching Decoder.

\subsubsection{Encoder0}

\textbf{\emph{Encoder0}} is the first part of FE encoder that does supervised learning. If looked separately, it’s exactly the same as a normal supervised learning network. The purpose of \textbf{\emph{Encoder0}} is for extracting labels (human-selected representations) from the data. When there’s no label available, it can also work without supervision to extract the “Principal latent variables” of the data.
The input of \textbf{\emph{Encoder0}} is the data vector $x$, and the output is either the predicted label vector $\hat{y}$ or the principal latent vector $z_{0}$ with 1 column.

When used with supervision, the objective of \textbf{\emph{Encoder0}} is to minimize the difference between actual label $y$ and output $\hat{y}$. With loss function:

\begin{equation}
\mathcal{L}_{Encoder_{0}} = \mathbb{E}_{x,y\sim Pd}(y-\hat{y})^2
\end{equation}

where $\hat{y} = Encoder_{0}(x)$. With continuous labels, Mean Squared Error (MSE) is used to calculate the error between the predicted and actual label, for categorical labels, Cross Entropy can be used instead.

When \textbf{\emph{Encoder0}} is used without supervision, the objective of \textbf{\emph{Encoder0}} is of the same kind with the \textbf{\emph{Encoder0}} and will be discussed together in the \textbf{\emph{Encoder0}} Session.

\subsubsection{Encoder}

The \textbf{\emph{Encoder}} of FE is like the VAE Encoder, it learns to output a distribution as latent representations from the input data $x$. The final output of \textbf{\emph{Encoder}}, $z$, is sampled from the Gaussian distribution $\mathcal{N}(\mu,\sigma^2)$ with parameters $\mu$ and $\sigma$ using the reparameterization trick proposed in VAE. $\mu$ and $\sigma$ are the outputs of the \textbf{\emph{Encoder}}:

\begin{equation}
z=\mu+\sigma\cdot\epsilon
\end{equation}

where $\epsilon\sim \ \mathcal{N}\left(0,I\right)$. The output $z$ has $n$ columns, $n$ is defined according to the task requirements. For each column of $z$, a reconstruction result $\hat{x_{i}}$ is generated by the PPD, and the \textbf{\emph{Encoder}}’s goal is to minimize the differences between each $\hat{x_{i}}$ and input data $x$. A hyperparameter $\xi$ is involved in the loss function to balance the optimization process, and to put equal effort on each reconstruction result – that the reconstruction error weighted by $\xi$ should be roughly the same. According to experiment results, we decided that using a geometric series is appropriate, $\alpha$ is the common ratio of the geometric series, normally $\alpha=\frac{2}{3}$. Therefore, the reconstruction loss term of Encoder loss function is:

\begin{equation}
\mathcal{L}_{Encoder_{RL}}=\mathbb{E}_{x\sim Pd}\left\{\sum_{i=1}^{n}\left[\left(x-{\hat{x}}_i\right)^2\cdot\left(\xi\cdot\frac{1-\alpha^i}{1-\alpha}+1\right)\right]\right\}
\end{equation}

where ${\hat{x}}_i={PPD\left(Encoder\left(x\right)\right)}_i$. The trend of multiplier $\left(\xi\cdot\frac{1-\alpha^i}{1-\alpha}+1\right)$ is a geometric series.

Adding the KL divergence term that encourages the posterior $q\left(z\middle| x\right)$ to be as close to the Gaussian prior $p\left(z\right)$, the complete \textbf{\emph{Encoder}} loss function is:

\begin{equation}
\mathcal{L}_{Encoder}=\mathcal{L}_{Encoder_{RL}}-\frac{1}{n}\cdot \sum_{i=1}^{n}D_{KL}[q(z_{i}|x)||p(z_{i})]
\end{equation}

For unsupervised \textbf{\emph{Encoder0}}, the purpose is to find the first principal representation. It only needs to minimize the reconstruction error of $\hat{x}_{0}$, equivalent to a traditional VAE with 1 latent variable. The loss function is therefore similar to which of the \textbf{\emph{Encoder}}:

\begin{equation}
\mathcal{L}_{Encoder_0}=\mathbb{E}_{x\sim Pd}\left[\left(x-{\hat{x}}_0\right)^2\right]- D_{KL}[q(z_{0}|x)||p(z_{0})]
\end{equation}

where ${\hat{x}}_0={PPD\left(Encoder\left(x\right)\right)}_0$. Because they share common structures, the 2 loss functions can be combined for easier programming in practice.

To improve the robustness of the feature extractor \textbf{\emph{Encoder0}} and \textbf{\emph{Encoder}}, noise is added to the data $x$ before it’s fed into the neural nets. A dropout layer is used in this case. This special layer can set a certain amount of network inputs to 0, forcing the encoder nets to extract representations with fewer observable variables. This operation can also be regarded as the kind of self-supervised learning method that mask some random variables and try to reconstruct them with the other variables.

\subsection{Progressive Patching Decoder (PPD)}

The latent vector $z$ is generated one column at a time by the encoder network of FE, each of them contributing a new degree of freedom to the reconstruction result. PPD takes in the n-dimensional $z$ and produces n results using 1,2,…,$n$ latent variables. PPD is designed to be a multi-level structure where the newly added latent variable is used to “patch” the previous result and refine it by giving more information.

From a regular feature fusion point of view, the “patching” process should either be done by concatenating the vectors or adding them together. However, we cannot add the latent variables together directly, as it would limit the width of AE bottlenecks and provide fewer degrees of freedom. Concatenating also doesn’t work in this case, as the input shape for a neural network layer should normally be fixed, the structure with adaptive shapes and elements from the continual learning domain might work, but is unnecessarily difficult for this task.

Achieving both utility and simplicity, PPD uses a hierarchical structure: the first latent variable is converted by a neural network NN0 into the base median vector $m_{0}$, the newly added latent variables are converted by NNi into the patcher vector $p_{i}$ and “patch” $m_{0}$ one by one, then both the patched and under-patched median vectors are used for reconstruction. The median vectors should have dimensions bigger than n, so they can include all information from latent variables. The patching process can be done by simply adding the patcher vector to the median vector: 

\begin{equation}
m_i=p_i+m_{i-1}
\end{equation}

As we discovered later, a 2-step linear transformation achieves the same result and converges faster:

\begin{equation}
m_i=p_{i_1}+p_{i_2}\cdot m_{i-1}
\end{equation}

where $p_{i_1}$ is the adder patcher vector and $p_{i_2}$ is the multiplier patcher vector, both are generated by $NN_i$ and have the same shape with $m_i$. $p_{i_1}$ is initialized to be a gaussian distribution with 0-mean and $p_{i_2}$ to have 1-mean.
Originally, the median vector m is designed to be n-dimensional, acquired by padding zeros around the 1-dimensional latent vector. Adding the median vectors together is then equivalent to concatenating the latent vectors. However, this process puts extra pressure to the decoder, reducing overall efficiency. The current version relaxed the restrictions and allowed $m$ to be generated more freely, also enabling the multiplier patch from Eq.7 to work.

\subsubsection{NN0}

$NN_0$ is the first part of Progressive Patching Decoder. Its purpose is to take in $z_0$ generated by \textbf{\emph{Encoder0}} and output the base Median Code $m_0$. If $z_0$ is the predicted label, we can also use the true label as input of $NN_0$ while training to speed up the training process.

The objective of $NN_0$ is to minimize the reconstruction loss of $\hat{x}_0$, which is calculated solely by $m_0$. The loss function is:

\begin{equation}
\mathcal{L}_{{NN}_0}=\mathbb{E}_{x\sim Pd}\left[\left(x-{\hat{x}}_0\right)^2\right]
\end{equation}

\subsubsection{NNi}

$NN_i$ generates the progressive patcher for median code to calculate the refined median codes. The total amount of $NN_i$ is decided according to the task. The input for $NN_i$ is $z_i$ generated by \textbf{\emph{Encoder}}, and the output is $p_{i1}$ and $p_{i2}$ the adder and multiplier patcher.

The objective of $NN_i$ is to minimize the reconstruction loss of ${\hat{x}}_i$, which is calculated by $m_0$,\ $m_1$, ...~,$\ m_i$. The loss function is:

\begin{equation}
\mathcal{L}_{{NN}_i}=\mathbb{E}_{x\sim Pd}\left[\left(x-{\hat{x}}_i\right)^2\right]
\end{equation}

\subsubsection{Decoder}

The Decoder of PPD is the hardest to train, its goal is to turn every $m$ into $\hat{x}$ with the highest accuracy possible. It is pushed to find the general mapping between the \emph{median code} and \emph{reconstruction result}, and this mapping should work regradless if the information is enough. The properly trained FE \textbf{\emph{Decoder}} is like when decoders from many trained autoencoders with different numbers of latent variables are squeezed together.

The loss function of Decoder is very similar to $\mathcal{L}_{Encoder_{RL}}$, with $\hat{x}_0$ included:

\begin{equation}
\mathcal{L}_{Decoder}=\mathbb{E}_{x\sim Pd}\left\{\sum_{i=0}^{n}\left[\left(x-{\hat{x}}_i\right)^2\cdot\left(\xi\cdot\frac{1-\alpha^i}{1-\alpha}+1\right)\right]\right\}
\end{equation}

\section{Experiment}
\label{Exp}

To assess the quality of the proposed Full Encoder network, we conducted experiments on a toy dataset sampled from a custom-made non-linear system. Results on FE with different settings are collected and compared to the more classical VAE and beta-VAE framework. Mutual Information [16] is used to measure the relativity of the representations extracted and generative factors.

We also did some supplementary experiments to further explore the capabilities of Full Encoder. The first experiment presents FE being applied to a linear toy dataset to compare results with the PCA method. The second experiment presents FE’s power to learn with labels first then extract the latent representations (semi-supervised). The third experiment presents FE’s performance on the more complicated MNIST dataset, showing promise in extracting stable variables from more complicated real-life datasets. Figures for more experiments please see Appendix B.

\subsection{Non-linear Toy Dataset}

We want to present the full potential of Full Encoder in a clear yet not too easy or challenging way. Therefore, we decided to design a non-linear toy dataset and test our model on this dataset first, before moving to any real-world datasets. The toy datasets are easy to control and allow us to discover some interesting properties of Full Encoder, while presenting decent challenge and complexity.

For the dataset, we first created a mathematical model for the custom-made non-linear memoryless time-invariant system, and use the model to generate the data. Fig.~3 shows the degree of non-linearity the mappings used for the costume-made system. The system has 5 inputs (generative factors) and 48 outputs (observable variables), each output is determined by the 5 inputs under a random non-linear transformation with Gaussian noise, see Fig.~3.

Each input from the toy dataset is not equivalent to each other, as the “importance” of each input variable is different, determined by its overall contribution to the outputs. A less “important” input variable would have a smaller impact on output variables with the same degree of changes, and the non-linearity of the impact is also reduced. In our toy datasets, the 1st and 2nd input variables (generative factors) $S1$, $S2$ have equal importance and $S3\sim S5$ have gradually decreased importance.
The dataset is then generated from the model with 5 inputs independently sampled from truncated Gaussian distributions with standard deviation of 1. 

\begin{figure}
  \centering
  \includegraphics[width=0.75\textwidth]{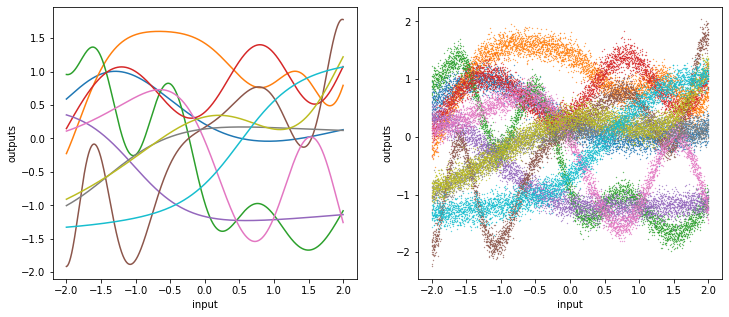}
  \caption{Figure shows a few systems outputs changing accordingly to one input varying from -2 to 2 while other inputs are fixed to 0. The figure on the left shows the non-linear mapping without noise, the figure on the right is what the dataset truly looks like when adding Gaussian noise with a standard deviation of 0.125.}
\end{figure}

\subsection{Experiment Settings}

The architecture of Full Encoder used in the experiment session for the toy dataset is shown in Table~1. All experiments including those from other frameworks (VAE, etc.) are conducted on this architecture, with different $NN_i$ unit amounts.

\begin{table}
  \caption{Full Encoder architecture for Non-linear Toy datasets}
  \label{sample-table}
  \centering
  \begin{tabular}{lll}
    \toprule
	\textbf{\emph{Encoder0}} and \textbf{\emph{Encoder}} & \textbf{\emph{NNi}} &\textbf{\emph{Decoder}} \\
    \midrule
    Input 48 1d array                          & Input $z_i\in\mathbb{R}^1$              & Input $m_i\in\mathbb{R}^50$     \\
    Dropout (Drop Ratio: 0.2)                  &                                         &              \\
    FC. 64 SeLU                                & FC. 32 SeLU                             & FC. 64 SeLU  \\
	FC. 64 SeLU                                & FC. 32 SeLU                             & FC. 64 SeLU  \\
	FC. $2\cdot n^* (\mu_i,\ \sigma_i)$     & FC. $2\cdot 50 (p_{i1},\ p_{i2})$  & FC. 48       \\
    \bottomrule
	\multicolumn{2}{c}{$n^*$ is determined according to the experiment requirements}
	
  \end{tabular}
\end{table}

While presenting the experiment results we are using $L1\sim Ln$ to represent the $n^th$ latent variable, $z_i$ from the architecture is not used here, as $z_0$ sometimes represent more than 1 latent variable and may be confusing, to make it clearer we used Li to present the latent variables. $S1\sim S5$ are used to represent the generative factors or the inputs of the dataset respectively.

A total of 9 experiments is conducted to present the properties of FE with the toy dataset. In the first 6 experiments, the number of latent variables given to FE is increased by 1 each time; the 7th experiment used a standard VAE model with 6 latent variables; the 8th experiment used a corresponding beta-VAE with 6 latent variables; in the 9th experiment, we combined the mechanisms used in beta-VAE and FE and tests the result of $\beta$-FE with 6 latent variables.

In each experiment, 10000 data pieces are sampled from the toy dataset and a total of 20000 training iterations are conducted (whether it’s converged or not) using Adam optimizer with batchsize = 500 and learning rate = 0.001. Note that each experiment presents in this section is done individually, they have different network random initial states and do not share any trainable parameters with each other. Each experiment is done twice with different initial states to prove reproduction stability of the proposed method.

\subsection{Experiment Results}

\subsubsection{Stability of Representations}

As shown in Fig. 3c and Fig. B1c$\sim$Fig. B5c (from Appendix~B), the non-linear mapping that presents the relationship between $S$'s (generative variables) and $L$'s (latent variables) shows the representations learnt by each $L$'s of the Full Encoder model. Through the patterns of the mini-figures, though messy as they may look like, great consistency can be observed. The 2 results from the same experiment setting but with differnt initializations are always similar, meaning the method have very high reproduction ability. The $L_i$ from Full Encoder with 1 Latent variable and all the $L_i$'s ($i = 1,2,...,5$) from FE with $2\sim$6 latent variables are also learning basically the same pattern, meaning the stability is still effective with different hyperparameter settings. We also tried changing the activation function, it didn't change the observation as well. Comparing to VAE and $\beta$-VAE as shown in Fig. B7 and Fig. B8, methods without the Progressive Patching Decoder can't regulate which latent variable to learn which representation. From Fig. B9 we see $\beta$-VAE when combined with FE can gain stability well too, with some minor negative effect on representation interpretability.

We can see from this part that representations learnt by the Full Encoder show strong stability and consistency, and can easily be reproduced -- even when the network structure is different. The representations learnt by FE, given enough training iterations, network capacity and a constant dataset, should always converge at a fixed point, rather a golbal minimum than a local one. 

\begin{figure}
  \centering
  \includegraphics[width=0.85\textwidth]{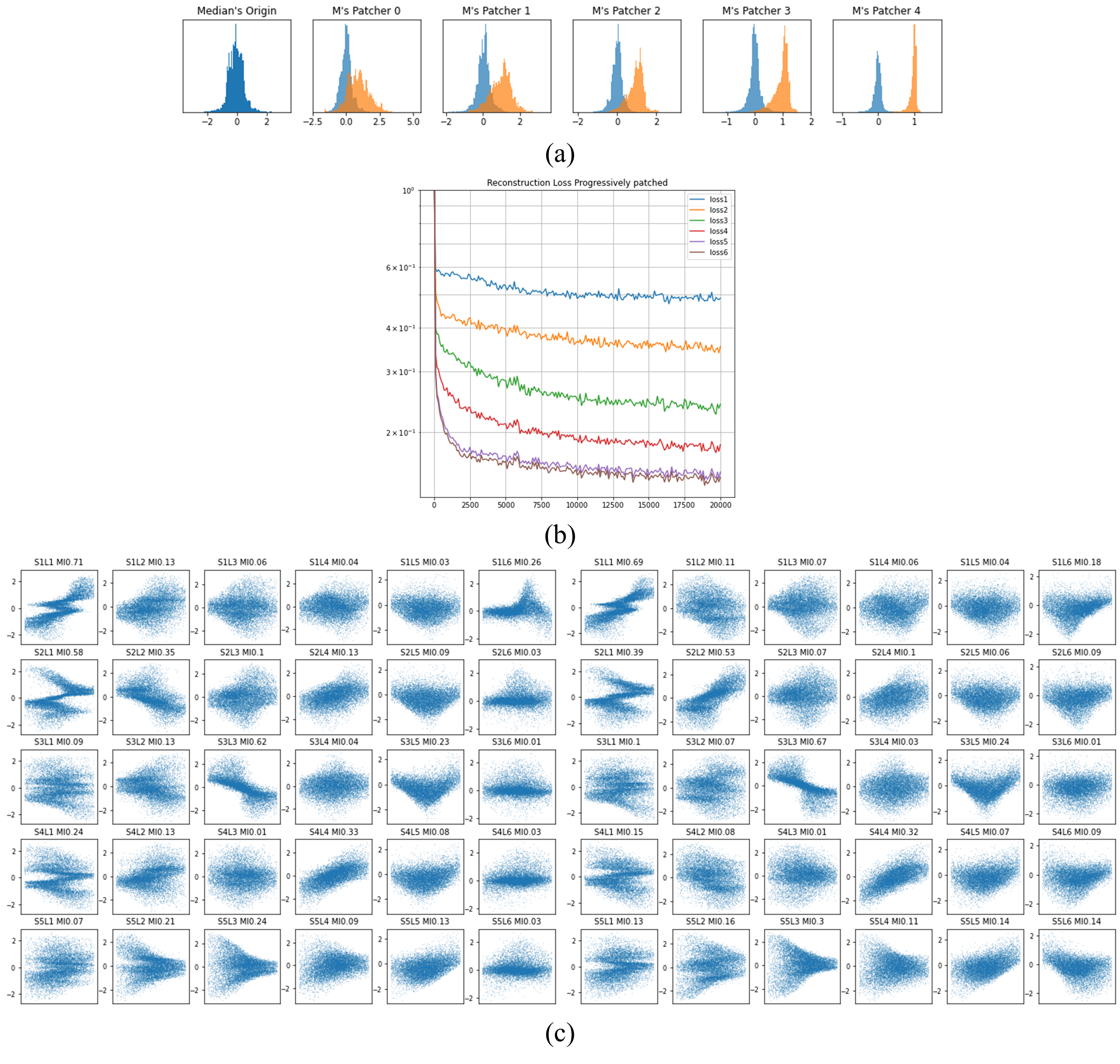}
  \caption{(a) shows the histogram of patcher codes and median codes. As levels of PPD patchers get higher, the patchers are getting closer to center, meaning they are providing less new and significient information for reconstruction. (b) shows the reconstruction loss achieved by first $i$ latent variable. Each new latent variables added lower the reconstruction error by a certain amount, until all degrees of freedom are taken. (c) shows the non-linear mapping between L1$\sim$L6 and S1$\sim$S5, the patterns formed are used to identify what representations the latent variables are learning. Patterns on both sides are from different experiments, yet they look very alike, showing FE's power to maintain a unified result regardless of network initializations.}
\end{figure}

\subsubsection{Progressive Patching and Degree of Freedom of the System}

As shown in Fig. 3a and Fig. B1a$\sim$Fig. B5a (from Appendix~B), the patcher codes are gradually closing in to center, meaning they are contributing less and less to the final result. This is also reflected in Fig. 3b and Fig. B1b$\sim$Fig. B5b (from Appendix~B), the reconstruction error is gradually decreasing with more latent variables added and at a certain level, the decrease seemd to cease.

Experiments of FE with 2$\sim$5 latent variables have shown FE progressively refining the reconstruction results, exploring new degrees of freedom for refinement while leaving the patterns learned by the previous latent variables intact. Because the system has only 5 generative factors, when given the 6th latent variable, it become redundant can couldn't refine the result any further. The first 5 latent variables L1$\sim$L5 covered up all the degrees of freedom provided by the 5 generative factors in the system.

Table 2 shows the reconstruction error of all the FE experiments along with VAE, $\beta$-VAE, $\beta$-FE and supervised FE, with different latent variables available to use. The training stops at 20000 iterations so the neural nets were not fully converged, we deliberately designed this so the difference between the converging speed can be shown as well, because adding new latent variables will not affect the reconstruction error achieved by the older latent variables, therefore when fully converged, the colomns of the first 6 rows should all be the same.

\begin{table}
  \caption{Reconstruction Errors for Different Architectures with Different Amount of Latent Variables}
  \label{sample-table}
  \centering
  \begin{tabular}{lllllll}
    \toprule
	 - & R.E L1 & R.E L2 & R.E L3 & R.E L4 & R.E L5 & R.E L6 \\
    \midrule
	FE - 1 L.V & 0.422 & - & - & - & - & - \\
	FE - 2 L.V & 0.431 & 0.305 & - & - & - & - \\
	FE - 3 L.V & 0.455 & 0.308 & 0.235 & - & - & - \\
	FE - 4 L.V & 0.448 & 0.315 & 0.226 & 0.188 & - & - \\
	FE - 5 L.V & 0.472 & 0.335 & 0.231 & 0.183 & 0.161 & - \\
	FE - 6 L.V & 0.484 & 0.348 & 0.235 & 0.179 & 0.149 & 0.147 \\
	VAE - 6 L.V & - & - & - & - & - & \bf0.122 \\
	$\beta$-VAE - 6 L.V & - & - & - & - & - & 0.155 \\
	$\beta$-FE - 6 L.V & 0.565 & 0.426 & 0.331 & 0.253 & 0.186 & 0.183 \\
	Supervised-FE & - & 0.450 & 0.281 & 0.194 & 0.143 & 0.141 \\
    \bottomrule
	
  \end{tabular}
\end{table}

It is worth noticing that normal VAE out-performs all the other architectures with same amount of latent variables reconstruction error-wise, it's beating $\beta$-VAE and $\beta$-FE because the disentanglement mechanism is achieved by limiting the information bottleneck in this branch, so while reconstructing $\beta$-VAE lose more information than VAE. The other FE models are out-performed by VAE because they are searching for a global minimum rather than a local minimum, making the convergence process much longer.

\subsubsection{Semi-supervised FE on MNIST}

For the MNIST dataset, Semi-supervised FE is used to trained with supervision to generate the digits, then learn the other properties of these hand-written digits by itself. The first 3 properties learned are fixed and human-interpretable: Angle, Width, and Thickness. The other properties learned are not so easily described. These 3 factors are selected by the Full Encoder as the most important factors for digits reconstruction from MNIST dataset. FE can also generate the digits using only several selected latent variables with different reconstruction fidelity, the “intermediate products” as they are refined to high quality pictures. Results are shown in Fig.~B12.

\subsubsection{Other Experiments}

Experiment show FE's flexibility to work with labelled data. Results see Fig. B11(Appendix B). Experiment show linear FE can achieve the same result as PCA. Results see Fig. B10 (Appendix B).

\section{Discussion}

Inspired by PCA and the $\beta$-VAE branch, Full Encoder is proposed in this paper to provide a new horizon for training an autoencoder. To control what the unsupervised deep learning latent variable methods are learning and try to make them interpretable has been very hard. Full Encoder managed to stabilize the latent variables and forced a specific latent variable to learn the representation by the rank of importance. When combined with $\beta$-VAE, the Full Encoder can further be used to analyze the importance of generative factors for reconstruction.

Punishing the latent variables hard grants FE many useful properties, but also results in difficulty of training. FE doesn’t cope with large systems with hundreds of generative factors well, as the latent variables required might be too many and would make the graph too complicated and take much longer to converge than a normal VAE model. The raw FE model also generate representatioins with less human-interpretability, as importance for reconstruction and interpretability can often be conflictive. FE can only work with memoryless systems right now and doesn’t include any time series non-linear analysis, which can be inconvenient for some industrial applications.

To also try to fit in the bigger picture, from the blog [14] Yann LeCun and Ishan Misra wrote in Chapter “Non-contrastive energy-based SSL” that they are looking for “a method to minimize the capacity of the latent variable”. The mechanism proposed in Full Encoder can do exactly that, by limiting the number of latent variables, indirectly controlling what is represented. However, the raw form of Full Encoder hasn’t any chance applied to high resolutions vision or NLP tasks, more enhancements are to be done to make this architecture more useful.

Overall, Full Encoder is a flexible architecture that can be used in simple real-life scenarios and would help us analyze non-linear systems better. By simply adjusting the parameters of Full Encoder, it can be turned into a regular AE, a conditional AE, a VAE, a beta-VAE, a denoising AE or a sparse AE, along with the new potentials brought with the Progressive Patching Decoder; it can be used with or without supervision, get disentangled interpretable representations, acquire robust and stable latent variables, and figure out the degrees of freedom of a system.

It includes in many of the classical autoencoders organically and could be a useful tool for small-scale real-life industrial problems, therefore given the name of Full Encoder.

\begin{ack}
Thanks to a friend of mine who inspired me on the name "FE".
\end{ack}

\section*{References}

{
\small

[1]	Rifai, S., Vincent, P., Muller, X., Glorot, X.\ \& Bengio, Y.\ (2011). Contractive Auto-Encoders: Explicit Invariance During Feature Extraction. {\it ICML.}

[2]	Kingma, D.P.\ \& Welling, M.\ (2014). Auto-Encoding Variational Bayes. {\it CoRR, abs/1312.6114.}

[3]	Kingma, D.P., Mohamed, S., Rezende, D.J.\ \& Welling, M.\ (2014). Semi-supervised Learning with Deep Generative Models. {\it NeurIPS.}

[4]	Higgins, I., Matthey, L., Pal, A., Burgess, C., Glorot, X., Botvinick, M., Mohamed, S.\ \& Lerchner, A.\ (2017). beta-VAE: Learning Basic Visual Concepts with a Constrained Variational Framework. {\it ICLR.}

[5]	Burgess, C., Higgins, I., Pal, A., Matthey, L., Watters, N., Desjardins, G.\ \& Lerchner, A.\ (2018). Understanding disentangling in beta-VAE. {\it arXiv: Machine Learning.}

[6]	Chen, T.Q., Li, X., Grosse, R.B.\ \& Duvenaud, D.\ (2018). Isolating Sources of Disentanglement in Variational Autoencoders. {\it NeurIPS.}

[7]	Esmaeili, B., Wu, H., Jain, S., Bozkurt, A., Siddharth, N., Paige, B., Brooks, D.H., Dy, J.\ \& Meent, J.V.\ (2019). Structured Disentangled Representations. {\it AISTATS.}

[8]	Goodfellow, I.J., Pouget-Abadie, J., Mirza, M., Xu, B., Warde-Farley, D., Ozair, S., Courville, A.C.\ \& Bengio, Y.\ (2014). Generative Adversarial Nets. {\it NeurIPS.}

[9]	Mirza, M.\ \& Osindero, S.\ (2014). Conditional Generative Adversarial Nets. {\it ArXiv, abs/1411.1784.}

[10]	Donahue, J., Krähenbühl, P.\ \& Darrell, T.\ (2017). Adversarial Feature Learning. {\it ArXiv, abs/1605.09782.}

[11]	Chen, X., Duan, Y., Houthooft, R., Schulman, J., Sutskever, I.\ \& Abbeel, P.\ (2016). InfoGAN: Interpretable Representation Learning by Information Maximizing Generative Adversarial Nets. {\it NeurIPS.}

[12]	Lin, Z., Thekumparampil, K.K., Fanti, G.\ \& Oh, S.\ (2020). InfoGAN-CR and ModelCentrality: Self-supervised Model Training and Selection for Disentangling GANs. {\it ICML.}

[13]	Bengio, Y., Yao, L., Alain, G.\ \& Vincent, P.\ (2013). Generalized Denoising Auto-Encoders as Generative Models. {\it NeurIPS.}

[14] LeCun, Y.\ \& Ishan, M.\ (2021). Self-supervised learning: The dark matter of intelligence. {\it ai.facecbook.cocm/blog/self-supervised-learning-the-dark-matter-of-intelligence/}

[15] LeCun, Y.\ \& Cortes, C.\ (2005). The mnist database of handwritten digits.

[16]	Kraskov, A., Stögbauer, H.\ \& Grassberger, P.\ (2004). Estimating mutual information. {\it Physical review. E, Statistical, nonlinear, and soft matter physics, 69 6 Pt 2,} 066138 .

}

\section*{Checklist}

\begin{enumerate}

\item For all authors...
\begin{enumerate}
  \item Do the main claims made in the abstract and introduction accurately reflect the paper's contributions and scope?
    \answerYes{}
  \item Did you describe the limitations of your work?
    \answerYes{}
  \item Did you discuss any potential negative societal impacts of your work?
    \answerNo{The work is a foundational research that focuses on understanding the non-linear systems and revealing the link between their inputs and outputs, the negative impacts are only associated to the systems someone might use this work on. It is currently too primitive to be applied to any system that could bring any societal impacts.}
  \item Have you read the ethics review guidelines and ensured that your paper conforms to them?
    \answerYes{}
\end{enumerate}

\item If you are including theoretical results...
\begin{enumerate}
  \item Did you state the full set of assumptions of all theoretical results?
    \answerNA{}
	\item Did you include complete proofs of all theoretical results?
    \answerNA{This work is mostly experiment based, because the theory behind it (as discussed in Appendix A.2 Methodology) seems rather simple.}
\end{enumerate}

\item If you ran experiments...
\begin{enumerate}
  \item Did you include the code, data, and instructions needed to reproduce the main experimental results (either in the supplemental material or as a URL)?
    \answerYes{The data files and the jupyter notebook file of which the experiments are conducted on is submitted as part of the supplemental material.}
  \item Did you specify all the training details (e.g., data splits, hyperparameters, how they were chosen)?
    \answerYes{}
	\item Did you report error bars (e.g., with respect to the random seed after running experiments multiple times)?
    \answerNo{This work doesn't care too much about error rates. However, regarding the random seeds of running experiments, we present 2 results with different random initializations for all experiments, to show the stability of our method.}
	\item Did you include the total amount of compute and the type of resources used (e.g., type of GPUs, internal cluster, or cloud provider)?
    \answerNo{The neural net required for this work is rather small, people can run it anywhere. In our case, we used a computer with one Nvidia RTX 2060 GPU. Training time should be between 5 mins to 2 hrs, depending on the settings on number of latent variables.}
\end{enumerate}

\item If you are using existing assets (e.g., code, data, models) or curating/releasing new assets...
\begin{enumerate}
  \item If your work uses existing assets, did you cite the creators?
    \answerYes{The digital number MNIST dataset is used and cited.}
  \item Did you mention the license of the assets?
    \answerNA{}
  \item Did you include any new assets either in the supplemental material or as a URL?
    \answerYes{The data and the codes are provided in the supplemental material.}
  \item Did you discuss whether and how consent was obtained from people whose data you're using/curating?
    \answerNA{}
  \item Did you discuss whether the data you are using/curating contains personally identifiable information or offensive content?
    \answerNA{}
\end{enumerate}

\item If you used crowdsourcing or conducted research with human subjects...
\begin{enumerate}
  \item Did you include the full text of instructions given to participants and screenshots, if applicable?
    \answerNA{}
  \item Did you describe any potential participant risks, with links to Institutional Review Board (IRB) approvals, if applicable?
    \answerNA{}
  \item Did you include the estimated hourly wage paid to participants and the total amount spent on participant compensation?
    \answerNA{}
\end{enumerate}

\end{enumerate}

\end{document}